\documentclass[letterpaper, 10 pt, conference]{ieeeconf}  

\IEEEoverridecommandlockouts                              

\usepackage{amsmath,amssymb,amsfonts}
\usepackage{algorithmic}
\usepackage{graphicx}
\graphicspath{{./images}}
\usepackage{orcidlink}
\usepackage{todonotes}
\usepackage{booktabs}
\usepackage{subcaption}
\usepackage{lipsum}
\usepackage{paralist}
\usepackage{siunitx}

\let\labelindent\relax
\usepackage[inline]{enumitem}

\usepackage[backend=biber,style=ieee,minbibnames=1,maxbibnames=3]{biblatex}
\AtBeginBibliography{\footnotesize}
\usepackage[capitalize]{cleveref}
\crefname{section}{Sec.}{sections}
\Crefname{section}{Section}{Sections}
\crefname{lstlisting}{Lst.}{listings}
\Crefname{lstlisting}{Listing}{Listings}
\crefname{figure}{Fig.}{figures}
\Crefname{figure}{Figure}{Figures}
\crefname{table}{Tab.}{tables}
\Crefname{table}{Table}{Tables}
\crefname{algorithm}{Alg.}{algorithms}
\Crefname{algorithm}{Algorithm}{Algorithms}

\newcommand{\mm}{metamodel}
\newcommand{\mms}{\mm s}

\usepackage{xcolor}
\definecolor{codegray}{rgb}{0.5,0.5,0.5}
\definecolor{codegreen}{rgb}{0,0.6,0}
\definecolor{codebg}{rgb}{0.95,0.95,0.92}
\definecolor{codeblue}{RGB}{20,105,176}
\definecolor{codemagenta}{HTML}{F531D4}
\definecolor{codepurple}{HTML}{AD2AFE}
\definecolor{codeorange}{HTML}{E74D09}
\definecolor{codeteal}{HTML}{00CC9C}
\definecolor{mmred}{HTML}{B85450}
\colorlet{numb}{magenta!60!black}

\definecolor{mmscenario}{HTML}{E51400}
\definecolor{mmscrtmpl}{HTML}{FA6800}
\definecolor{mmscrvariant}{HTML}{D80073}
\definecolor{mmus}{HTML}{A20025}
\definecolor{mmvar}{HTML}{9C8300}
\definecolor{mmfluent}{HTML}{6A00FF}
\definecolor{mmtc}{HTML}{1BA1E2}
\definecolor{mmbhv}{HTML}{76608A}
\definecolor{mmtask}{HTML}{E3C800}
\definecolor{mmscene}{HTML}{60A917}
\definecolor{mmenvagn}{HTML}{6D8764}

\newcommand{\keyword}[1]{\texttt{\textbf{#1}}}

\newcommand{\libbehave}[1]{{\texttt{behave}}}
\newcommand{\libtextx}[1]{\texttt{textX}}
\newcommand{\libbddtextx}[1]{RobBDD}

\usepackage{listings}
\lstdefinelanguage{Gherkin}{
    morekeywords = {
        Given,
        When,
        Then,
        And,
        Feature,
        But,
        Background,
        Outline,
        Scenario,
        Outline,
        Examples,
        As,
        I,Want,
        So,That,
    },
    sensitive=true,
    morecomment=[l]{\#},
    commentstyle=\color{codegreen},
    morestring=[b]",
    breakatwhitespace=false,
    showstringspaces=false,
    breaklines=true,
    basicstyle=\ttfamily\scriptsize,
    numbers=left,
    numbersep=2pt,
    numberstyle=\tiny\color{codegray},
    backgroundcolor=\color{codebg},
    captionpos=b,
}

\lstdefinelanguage{textxbdd}{
morekeywords = {
    import,
    Behaviour,
    Event,
    Given,
    When,
    Then,
    And,
    User,
    Story,
    Scenario,
    Variant,
    Scenarios,
    Template,
    As,
    I, Want,
    So,That,
    var,
    template
    obj, ws, agn, event,
},
sensitive=true,
morecomment=[l]{\#},
commentstyle=\color{codegreen},
morestring=*[b]",
breakatwhitespace=false,
showstringspaces=false,
breaklines=true,
basicstyle=\ttfamily\scriptsize,
numbers=left,
numbersep=2pt,
numberstyle=\tiny\color{codegray},
backgroundcolor=\color{codebg},
captionpos=b,
literate=
    *{<}{{{\color{codeblue}{<}}}}{1}
    {>}{{{\color{codeblue}{>}}}}{1}
    {\{}{{{\color{codeblue}{\{}}}}{1}
    {\}}{{{\color{codeblue}{\}}}}}{1}
}

\lstdefinelanguage{json}{
    morekeywords = {
        @id,
        @type,
        @base,
        @context,
        @graph
    },
    sensitive=true,
    morestring=*[b]",
    breakatwhitespace=false,
    showstringspaces=false,
    breaklines=true,
    basicstyle=\ttfamily\scriptsize,
    numbers=left,
    numbersep=2pt,
    numberstyle=\tiny\color{codegray},
    backgroundcolor=\color{codebg},
    captionpos=b,
    literate=
    *{0}{{{\color{numb}0}}}{1}
    {1}{{{\color{numb}1}}}{1}
    {2}{{{\color{numb}2}}}{1}
    {3}{{{\color{numb}3}}}{1}
    {4}{{{\color{numb}4}}}{1}
    {5}{{{\color{numb}5}}}{1}
    {6}{{{\color{numb}6}}}{1}
    {7}{{{\color{numb}7}}}{1}
    {8}{{{\color{numb}8}}}{1}
    {9}{{{\color{numb}9}}}{1}
    {:}{{{\color{codeblue}{:}}}}{1}
    {,}{{{\color{codeblue}{,}}}}{1}
    {\{}{{{\color{codeblue}{\{}}}}{1}
    {\}}{{{\color{codeblue}{\}}}}}{1}
    {[}{{{\color{codeblue}{[}}}}{1}
    {]}{{{\color{codeblue}{]}}}}{1},
}

\lstdefinelanguage{turtle}{
    morekeywords = {
        @prefix, a
    },
    sensitive=true,
    morestring=[s][]{<}{>},
    breakatwhitespace=false,
    showstringspaces=false,
    breaklines=true,
    basicstyle=\ttfamily\scriptsize,
    numbers=left,
    numbersep=2pt,
    numberstyle=\tiny\color{codegray},
    backgroundcolor=\color{codebg},
    captionpos=b,
    literate=
    *{0}{{{\color{numb}0}}}{1}
    {1}{{{\color{numb}1}}}{1}
    {2}{{{\color{numb}2}}}{1}
    {3}{{{\color{numb}3}}}{1}
    {4}{{{\color{numb}4}}}{1}
    {5}{{{\color{numb}5}}}{1}
    {6}{{{\color{numb}6}}}{1}
    {7}{{{\color{numb}7}}}{1}
    {8}{{{\color{numb}8}}}{1}
    {9}{{{\color{numb}9}}}{1}
    {:}{{{\color{codeblue}{:}}}}{1}
    {,}{{{\color{codeblue}{,}}}}{1}
    {;}{{{\color{codeblue}{;}}}}{1}
    {\{}{{{\color{codeblue}{\{}}}}{1}
    {\}}{{{\color{codeblue}{\}}}}}{1}
    {[}{{{\color{codeblue}{[}}}}{1}
    {]}{{{\color{codeblue}{]}}}}{1},
}

\lstdefinelanguage{sparql}{
morekeywords = {
    PREFIX, CONSTRUCT, a, WHERE
},
sensitive=true,
morestring=[s][]{<}{>},
breakatwhitespace=false,
showstringspaces=false,
breaklines=true,
basicstyle=\ttfamily\scriptsize,
numbers=left,
numbersep=2pt,
numberstyle=\tiny\color{codegray},
backgroundcolor=\color{codebg},
captionpos=b,
literate=
*{:}{{{\color{codeblue}{:}}}}{1}
{?}{{{\color{codeblue}{?}}}}{1}
{;}{{{\color{codeblue}{;}}}}{1}
{.}{{{\color{codeblue}{.}}}}{1}
{\{}{{{\color{codeblue}{\{}}}}{1}
{\}}{{{\color{codeblue}{\}}}}}{1}
{[}{{{\color{codeblue}{[}}}}{1}
{]}{{{\color{codeblue}{]}}}}{1},
}

\lstdefinelanguage{jinja}{
morekeywords = {
    for, in, endfor, if
},
sensitive=true,
morestring=*[s][]{\{+}{+\}},
morestring=[s][]{\{\{}{\}\}},
morestring=*[b]",
breakatwhitespace=false,
showstringspaces=false,
breaklines=true,
basicstyle=\ttfamily\scriptsize,
stringstyle=\color{codepurple},
numbers=left,
numbersep=2pt,
numberstyle=\tiny\color{codegray},
backgroundcolor=\color{codebg},
captionpos=b,
literate=
*{.}{{{\color{codeblue}{.}}}}{1}
{\{}{{{\color{codeblue}{\{}}}}{1}
{\}}{{{\color{codeblue}{\}}}}}{1}
{\%}{{{\color{codeblue}{\%}}}}{1},
}

\lstdefinelanguage{console}{
morestring=*[b]",
sensitive=true,
breaklines=true,
showlines=true,
framesep=1pt,
basicstyle=\ttfamily\tiny,
stringstyle=\bfseries,
numberstyle=\tiny\color{codegray},
backgroundcolor=\color{codebg},
captionpos=b,
}

\newcommand{\inlinecode}[1]{\colorbox{codebg}{\texttt{#1}}}  

\title{\LARGE \bf
Automated Behaviour-Driven Acceptance Testing of Robotic Systems
}

\author{
Minh Nguyen$^{\ast \dagger}$~\orcidlink{0000-0002-0811-6441} %
\and
Sebastian Wrede$^{\dagger}$~\orcidlink{0000-0003-0029-8188} %
\and
Nico Hochgeschwender$^{\ast}$~\orcidlink{0000-0003-1306-7880} %
\thanks{$^{\ast}$~Faculty of Math. and Comp. Science, University of Bremen, Germany.{\newline} %
\texttt{\{minh.nguyen,nico.hochgeschwender\}@uni-bremen.de}} %
\thanks{$^{\dagger}$~Faculty of Technology, Bielefeld University, Germany.{\newline} %
\texttt{sebastian.wrede@uni-bielefeld.de}} %
\thanks{This work is supported in part by the European Union's H2020 project SESAME (Grant nr. 101017258) and the German Research Foundation (DFG), as part of Collaborative Research Center 1320 EASE – Everyday Activity Science and Engineering (\url{http://www.ease-crc.org/)}}
}

\begin{document}

\maketitle
\thispagestyle{empty}
\pagestyle{empty}

\begin{abstract}
The specification and validation of robotics applications require bridging the gap between formulating requirements and systematic testing. This often involves manual and error-prone tasks that become more complex as requirements, design, and implementation evolve. To address this challenge systematically, we propose extending behaviour-driven development (BDD) to define and verify acceptance criteria for robotic systems. In this context, we use domain-specific modelling and represent composable BDD models as knowledge graphs for robust querying and manipulation, facilitating the generation of executable testing models. A domain-specific language helps to efficiently specify robotic acceptance criteria.
We explore the potential for automated generation and execution of acceptance tests through a software architecture that integrates a BDD framework, Isaac Sim, and model transformations, focusing on acceptance criteria for pick-and-place applications.
We tested this architecture with an existing pick-and-place implementation and evaluated the execution results, which shows how this application behaves and fails differently when tested against variations of the agent and environment.
This research advances the rigorous and automated evaluation of robotic systems, contributing to their reliability and trustworthiness.

\end{abstract}

\section{Introduction} \label{sec:intro}
\begin{figure}[tp]
    \centering
    \begin{subfigure}[b]{0.49\linewidth}
        \includegraphics[height=0.145\textheight]{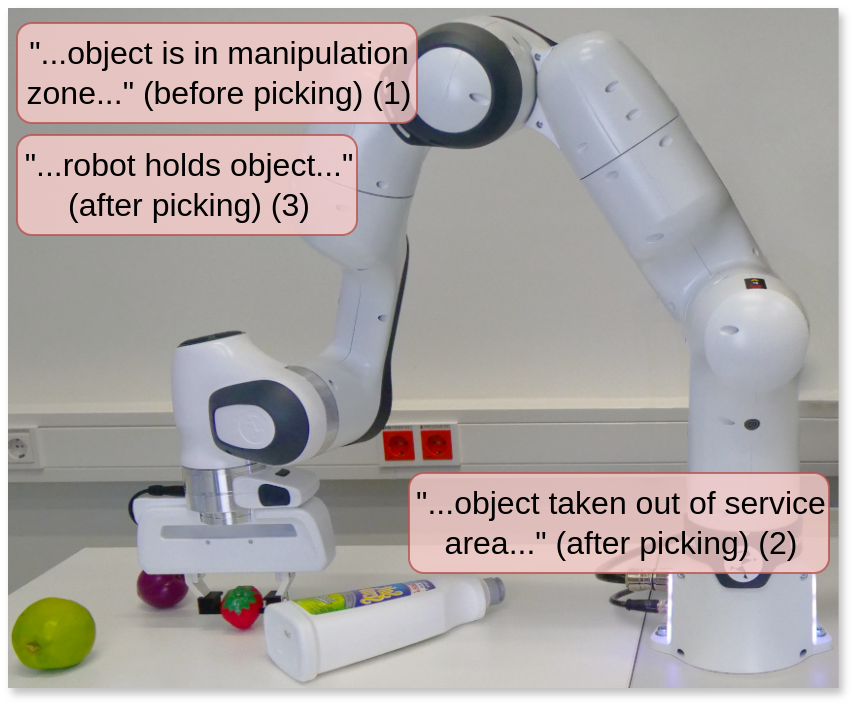}
        \subcaption{Picking from a service area}
        \label{fig:panda-fluent-pick}
    \end{subfigure}
    \hfill
    \begin{subfigure}[b]{0.49\linewidth}
    \includegraphics[height=0.145\textheight]{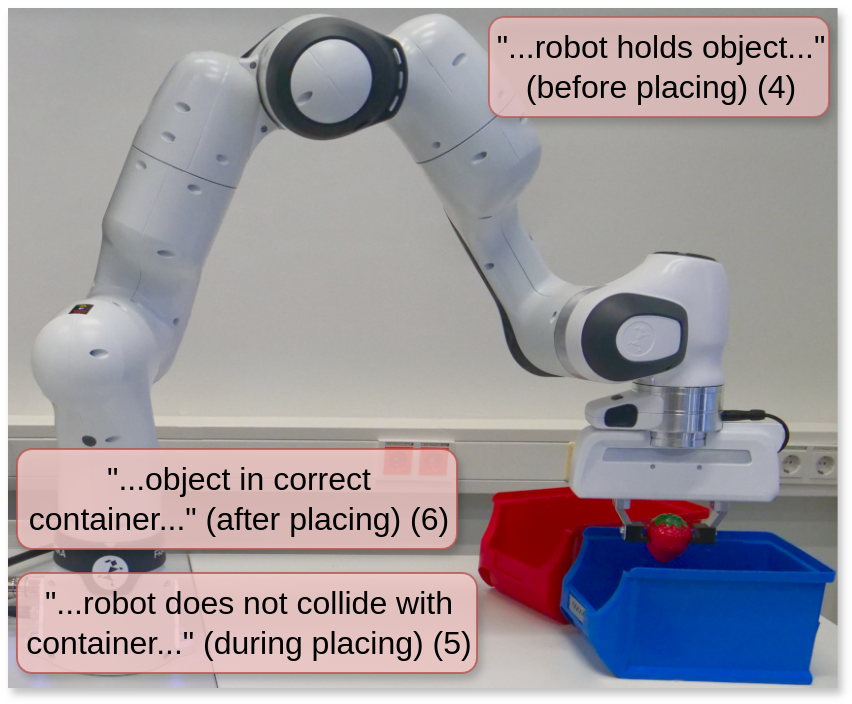}
    \subcaption{Placing in a container}
    \label{fig:panda-fluent-place}
    \end{subfigure}

    \caption{
        Sample acceptance criteria from the Basic Manipulation Test (e.g., pick-and-place task) in the RoboCup@Work~\cite{kraetzschmar2015} competition.
        Human judges assess compliance based on the rulebook, which details specifications for the manipulation zone, containers, etc.
    }
    \label{fig:panda-fluent}
\end{figure}

\noindent
The qualified deployment of robots necessitates verifying whether they meet their \emph{acceptance criteria (AC)} -- ensuring that their behaviour aligns with user requirements.
In software engineering, this is termed \emph{acceptance testing}~\cite{actesting}, a vital activity for high-quality system delivery.
In robotics, a special form of acceptance testing can be found in the field of performance evaluation and benchmarking through scientific competitions, such as RoboCup@Work~\cite{kraetzschmar2015}, RoCKIn~\cite{rockin}, and SOMA~\cite{mnyusiwalla2020}, where robot \emph{behaviours} are evaluated against their AC.
These assessments are typically outlined in benchmarking scenarios, rulebooks, or evaluation plans, often in ambiguous natural language.
Our prior work~\cite{nguyen2023} identified two common behaviours in these competitions, namely \emph{picking} and \emph{placing}\footnote{Without loss of generality, we will focus in this paper on pick-and-place behaviours, acknowledging that other behaviours such as navigation, pouring, and tool usage are also relevant.}, which are usually evaluated by human judges under varying environmental and task conditions.
Fig.~\ref{fig:panda-fluent} shows the pick and place behaviours and associated acceptance criteria from the RoboCup@Work rulebook~\cite{kraetzschmar2015}.

While scientific competitions can provide measures for the progress of a field and help identify areas for innovation, they are confined to carefully specified scenarios.
Resource constraints, such as human judges, lack of automation, and mere embedding of robots in real-world environments, usually allow only a subset of scenarios to be performed, which limits the validity of acceptance tests and resulting conclusions.
Motivated by our previous work~\cite{nguyen2023}, which identified common principles for benchmarking and evaluation among various competition protocols, we propose using \emph{behaviour-driven development} (BDD)~\cite{north2006bdd} as an effective method for defining and verifying robotic acceptance criteria.

BDD formulates AC as scenarios that capture the expected behaviours of the System under Test (SuT) using the construct: \keyword{Given} \texttt{<pre-condition>}, \keyword{When} \texttt{<action/event>}, \textbf{Then} \texttt{<expected outcome>}.
BDD frameworks like the Cucumber toolchain, which introduced the Gherkin syntax, support automatic executions for many programming languages.
Nevertheless, several conceptual deficits impede the direct application of these frameworks to test robotic acceptance criteria~\cite{nguyen2023}.
These include the absence of mechanisms to
\begin{inparaenum}[\itshape (i)\upshape]
    \item explicitly specify timing information about when to verify BDD clauses (\textbf{C1}),
    \item introduce domain-specific knowledge about the system, environment, and task, (\textbf{C2}) as well as
    \item express the relations between interdependent scenarios (\textbf{C3}).
\end{inparaenum}
We explore how our previous analysis~\cite{nguyen2023}, combined with domain-specific modelling
methods~\cite{nordmann2016}, can enhance the level of automation for acceptance testing in robotics. 
To this end, we investigate the following research questions:
\begin{enumerate}[label={\bfseries RQ\arabic*}]
    \item How can robotic acceptance criteria be represented to facilitate test automation?
    \item To what extent can the acceptance testing process be automated?
    \item How can automated test executions support the evaluation of robotic acceptance criteria?
\end{enumerate}
To address these questions, we make the following contributions that pave the way for automated acceptance testing of robotic systems:
\begin{inparaenum}[\itshape (i)\upshape]
    \item an extended \mm{} for BDD with enriched and new concepts required for specifying AC for robotics applications (\cref{sec:design}),
    \item a knowledge graph representation that enables querying and manipulation of AC, which ensures the validity of BDD models and supports the generation of execution artefacts (\cref{sec:dsl}),
    \item a domain-specific language that enables robot developers to specify AC models using the newly introduced concepts (\cref{sec:dsl}),
    \item an examination of the acceptance testing process to identify parts that can be automated, via a prototypical acceptance testing solution that leverages \libbehave{}, Isaac Sim, and model transformation (\cref{sec:exec}),
    \item and an evaluation of acceptance test results, demonstrating how our framework can provide evidence about the SuT's behaviour and failure causes in different scenario variations (\cref{sec:results}).
\end{inparaenum}
\cref{sec:discussion} continues with a discussion of the strengths and limitations of the approach presented, followed by a review of related work in \cref{sec:related} and a synopsis of the approach in \cref{sec:conclusion}.

\begin{lstlisting}[
float,
floatplacement=htbp,
language=Gherkin,
label=lst:example-gherkin,
caption={
    An excerpt of a BDD specification in the Gherkin syntax for the pick-and-place task.}
]
Feature: basic manipulation
  Background:
    Given a set of objects
    | ID    | Mass_g |
    | screw | 100    |
    | ...   | ...    |

  Scenario Outline: pick
    Given "<object>" is located at "<workspace>"
    When "<robot>" picks "<object>"
    Then "<object>" is held by "<robot>"
    And "<robot>" does not collide "<workspace>"
    Examples: Objects, workspaces, and robots
    | object | workspace | robot |
    | screw  | ws1       | panda |
    | ...    | ...       | ...   |

  Scenario Outline: place
  ...
\end{lstlisting}

\section{Conceptualizing Robotic Acceptance Criteria}
\label{sec:design}

\begin{figure*}[htbp]
    \centering
    \includegraphics[width=0.95\linewidth]{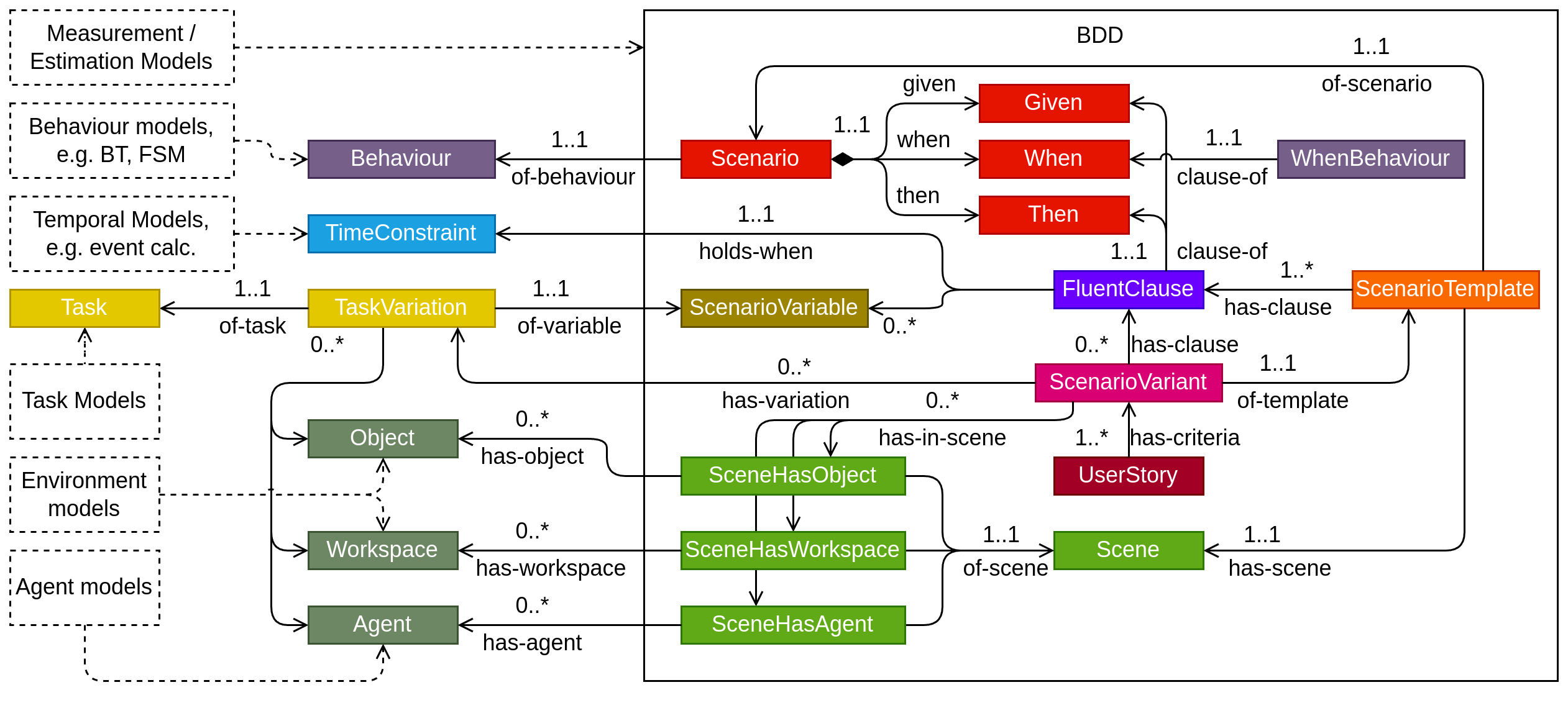}
    \caption{
        Overview of the concepts in our \mms{} and how they are linked to various aspects of a robotic scenario.
        Concretely defined concepts have solid borders, and arrows between these concepts conform to relationships in UML class diagrams.
        For the remainder of this paper, figures will use rounded rectangles to denote instances of these concepts, and figures and listings will use the same colour scheme to denote matching concepts and their instances.
    }
    \label{fig:bdd-overall}
\end{figure*}

\noindent
To address the conceptual deficits of existing BDD frameworks discussed in \cref{sec:intro}, we developed a \mm{} for BDD, detailed in this section.
\cref{fig:bdd-overall} shows how this \mm{} builds upon the core BDD structure -- a \keyword{Scenario} comprising \keyword{Given}, \keyword{When}, and \keyword{Then} concepts -- and links the introduced concepts to various robotic domains.

\begin{table}[htbp]
    \scriptsize
    \centering
    \caption{
        Example fluents matching the criteria in \cref{fig:panda-fluent}.
        The $t$-subscripts denote the fluents hold at different time.
    }
    \begin{tabular}{cp{0.8\linewidth}}
    \toprule
     Criterion & Matching fluent
    \\\toprule
    (1) & $holdsAt(locatedAt(object,manipulationZone), t_1)$
    \\\midrule
    (2) & $holdsAt(\neg locatedAt(object,serviceArea), t_2)$
    \\\midrule
    (3) & $holdsAt(isHeldBy(object,panda), t_2)$
    \\\midrule
    (4) & $holdsAt(isHeldBy(object,panda), t_3)$
    \\\midrule
    (5) & $holdsAt(\neg collides(object,container), t_4)$
    \\\midrule
    (6) & $holdsAt(locatedAt(object,container)\linebreak \wedge isCorrect(container), t_5)$
    \\\bottomrule
    \end{tabular}
    \label{tab:pickplace-panda-fluents}
\end{table}


\paragraph{BDD clauses}
Current BDD approaches generally do not support specifying \emph{when} scenario clauses should hold true (\textbf{C1}).
For instance, the ``is held by'' clause in \cref{lst:example-gherkin} asserts that the robot holds the target object \emph{after} completing the pickup behaviour.
The same assertion must hold \emph{before} the placing behaviour in the second scenario.
Such timing information is crucial in robotics, where interactions between robots and the environment are dynamic and complex.
Therefore, we model BDD clauses using the fluent concept, i.e. time varying properties of the world~\cite{miller2002}.
Fluents are used in several logic formalisms, e.g., event calculus~\cite{miller2002} and situation calculus~\cite{reiter1991,lakemeyer2010,thielscher2005}, to represent and reason about dynamical domains.
\cref{tab:pickplace-panda-fluents} shows how the criteria in \cref{fig:panda-fluent} can be represented using fluents in event calculus.
A fluent typically includes a predicate asserting a property of interest and a term indicating when the assertion should be valid.
These formalisms differ mainly in defining the latter term, e.g., time points in event calculus or abstract states in fluent calculus.

\begin{figure}[htbp]
    \centering
    \includegraphics[width=\linewidth]{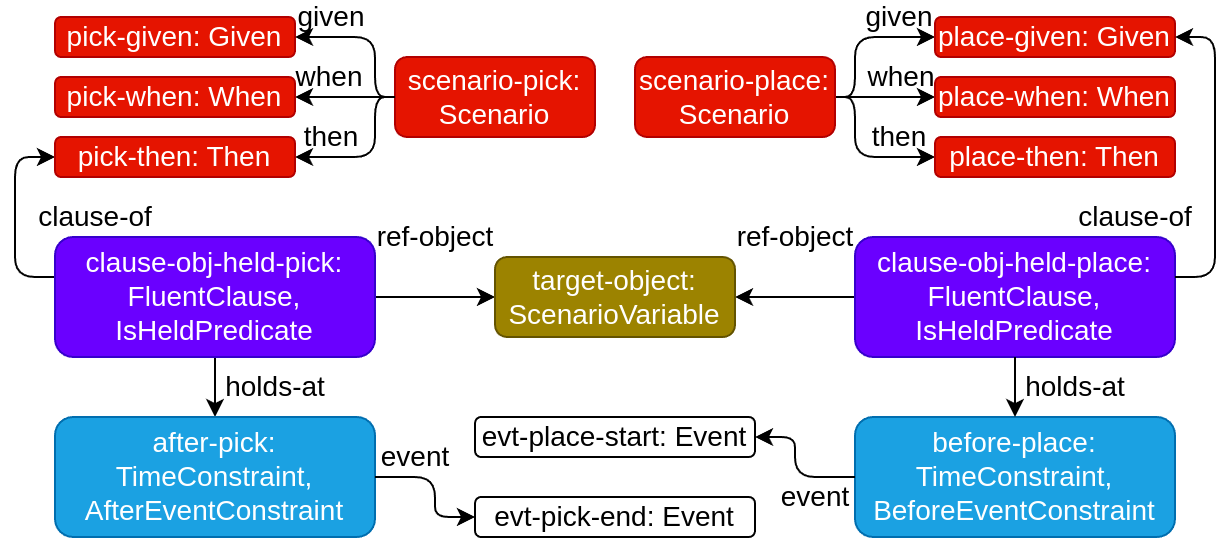}
    \caption{
        An example showing how the $isHeldBy$ fluent of criteria (3) and (4) in \cref{tab:pickplace-panda-fluents} can be represented with our \mm{} as two clauses in two separate BDD scenarios.
    }
    \label{fig:bdd-example-tmpl-fluent}
\end{figure}

In our models, a \keyword{FluentClause} can have more specific predicate types encoding semantics of the clause's assertion and handles two distinct compositions, i.e., relations between models, to represent a complete BDD clause.
\cref{fig:bdd-example-tmpl-fluent} illustrates this structure for the $isHeldBy$ fluent, e.g., criteria (3) and (4) in \cref{tab:pickplace-panda-fluents}.
First, the concept composes the fluent with its subjects, e.g., the object to be held.
We represent a predicate's subject as a \keyword{ScenarioVariable}, following logic terminology.
The variable's relation to the \keyword{FluentClause} instance denotes its role, e.g., \keyword{ref-object} in \cref{fig:bdd-example-tmpl-fluent} implies \texttt{target-object} is the object to be held.
Here, the variable concept expresses common elements across clauses and scenarios, e.g., the target object in a pick-and-place task.
The second composition links a \keyword{TimeConstraint} to the clause, indicating when the fluent should hold true, e.g., \texttt{after-pick} and \texttt{before-place} in \cref{fig:bdd-example-tmpl-fluent}.
These constraints reflect the distinct temporal semantics of the two clauses.
More specific types and compositions can extend the constraints to align with various formalisms for describing the system's temporal aspect.
For instance, in an event-based formalism, \texttt{after-pick} and \texttt{before-place} can be extended with types and compositions that tie them to events marking the end and beginning of respective behaviours, namely \texttt{after-pick-event} and \texttt{before-place-event} in \cref{fig:bdd-example-tmpl-fluent}.



\paragraph{Task variations and scene}
Our previous analysis~\cite{nguyen2023} identified objects, workspaces and agents as essential for specifying robotic acceptance criteria.
A typical BDD test would change these elements across different scenario variations, e.g., using Gherkin's \texttt{Examples} table as shown in \cref{lst:example-gherkin}.
To facilitate such variations, a \keyword{TaskVariation} in our \mm{} composes a \keyword{Task}, a \keyword{ScenarioVariable} and its possible variations, e.g., different objects to be picked.
The link to the variable, e.g., \texttt{target-obj} in \cref{fig:bdd-example-tmpl-fluent}, denotes what varies in the scenario.

Another mechanism connects a scenario to other elements anticipated during its execution but not explicitly mentioned by the task.
For instance, a scenario for selecting an object from a cluster may include assertions about one \emph{target object} but would require other objects to be present.
Gherkin partially supports this with the optional \texttt{Background} section in a feature file, interpreted as \texttt{Given} clauses for every scenario.
We, however, prefer to decouple these elements from the BDD clauses by introducing the \keyword{Scene} concept and reified relations linking a scene to objects, agents, and workspaces expected during scenario execution.



\paragraph{Domain-specific background knowledge}
The previously discussed concepts allow specifying BDD scenarios at an abstract level.
However, executing and evaluating variations of a scenario often requires switching between different domain-specific models of the changing object, workspace, or agent (\textbf{C2}).
For instance, to assert the $isHeldBy$ clause by checking the distance between the end-effector and the object after the pickup behaviour, geometric concepts are needed to represent this spatial relation.
Varying the object and agent requires updating this relation with the appropriate object coordinate frame and the agent's kinematics.
Such domain-specific knowledge can be introduced by adding types and compositions to agents, objects, and workspaces, akin to the time constraints shown in \cref{fig:bdd-example-tmpl-fluent}.
This link is decoupled from the scenario specification (unlike Gherkin's \texttt{Background}), allowing any clauses or scenarios linking to an object, agent, or workspace to utilize its domain-specific background knowledge, e.g., during test execution.


\paragraph{Scenario template, environment, and user stories}
Our \mm{} includes two composites to represent concrete BDD scenarios, namely \keyword{ScenarioTemplate} and \keyword{ScenarioVariant}.
A template integrates a \keyword{Scenario} with a scene and fluent clauses, while a variant combines a template with task variations and reified relations linking to expected scene elements, along with additional clauses needed to evaluate the concrete scenario.
Following the original BDD formulation~\cite{north2006bdd}, a \keyword{UserStory} (equivalent to \texttt{Feature} in Gherkin) is a composite of scenario variants.



\paragraph{Linking BDD scenarios to system behaviour models}
One limitation of existing BDD frameworks is their inability to manage interdependent scenarios (\textbf{C3}).
This is particularly pronounced in robotics, where behaviour-based coordination is essential to perform complex tasks.
For example, a pick-and-place solution may comprise two distinct phases to pick up and place the target object.
Here, a BDD scenario for the overall task is hierarchically related to its sub-behaviours, and the placing scenario depends on the successful completion of the picking phase.
Modelling the composition and coordination of behaviours is an enduring topic in robotics~\cite{nordmann2016} with numerous established methods such as state machines, behaviour trees~\cite{colledanchise2018}, and Petri nets~\cite{petri2008}.
We propose using coordination models from these formalisms to manage the relations between BDD scenarios by linking each \keyword{Scenario} to a corresponding execution unit in the coordination model, e.g., a state or action, through a \keyword{Behaviour}.
For example, a pick-and-place action node in a behaviour tree may have a sequence subtree with distinct action nodes for picking and placing.
Linking BDD scenarios to these nodes allows embedding the aforementioned hierarchical and sequential relations in the tree structure.
This can be achieved by introducing necessary types and compositions to a \keyword{Behaviour}, similar to the time constraints in \cref{fig:bdd-example-tmpl-fluent}.

%

\section{Representing Robotic Acceptance Criteria as Knowledge Graphs}\label{sec:dsl}
\noindent
We employ the established approach of domain-specific modelling and languages in robotics~\cite{nordmann2016} to represent a focused view of robotic acceptance testing.
To address \textbf{RQ1}, we formalize the concepts and their constraints in~\cref{sec:design} as knowledge graphs and implement a Domain-specific Language (DSL) and associated tooling for creating and transforming acceptance models into executable testing artefacts.



\begin{lstlisting}[
float,
floatplacement=htbp,
language=json,
label=lst:example-jsonld,
caption={
    JSON-LD model of a \keyword{FluentClause} linking to a scenario's \keyword{Then} element, a time constraint, and variables.
}]
{
  "@context":
  {
    "bdd": "https://my.url/metamodels/bdd#",
    %*{\color{mmfluent}"FluentClause"*): { "@id": %*{\color{mmfluent}"bdd:FluentClause"*) },
    "clause-of": {
        "@id": "bdd:clause-of", "@type": "@id" },
  },
  "@graph": [
  { "@id": %*{\color{mmscenario}"pickup-then"}*), "@type": "%*{\color{mmscenario}bdd:Then*)" },
  {
    "@id": %*{\color{mmfluent}"flc-held-pck"*),
    "@type": [ %*{\color{mmfluent}"bdd:FluentClause"*), "bdd:IsHeldPredicate"],
    "clause-of": %*{\color{mmscenario}"pickup-then"*),
    "holds-at": %*{\color{mmtc}"ftc-after-pick"*),
    "ref-object": %*{\color{mmvar}"var-target-obj"*),
    "ref-agent": %*{\color{mmvar}"var-pickplace-agn"*)
  } ]
}
\end{lstlisting}

\paragraph{Representing BDD models with JSON-LD}
The structure of our models and \mms{} is represented using JSON-LD\footnote{\url{https://www.w3.org/TR/json-ld/}}, a W3C standard extending JSON to encode linked data, enabling the utilization of the established tooling ecosystem surrounding JSON and Semantic Web standards.
JSON-LD introduces keywords that allow JSON objects to represent complex data compositions based on the Resource Description Framework (RDF)\footnote{\url{https://www.w3.org/TR/rdf11-concepts/}} data model.

\cref{lst:example-jsonld} shows a snippet of the JSON-LD representation for the $isHeldBy$ fluent clause in a pickup scenario, where keywords are prefixed with \inlinecode{@}.
The keywords most relevant for defining our \mms{} and models are \inlinecode{@context}, \inlinecode{@id}, and \inlinecode{@type}.
The context declares the vocabulary, i.e. \mm{}, for interpreting a JSON object, which allows models to conform to multiple domains by including relevant domain-specific concepts in the same context.
The \inlinecode{@id} keyword assigns an Internationalized Resource Identifier (IRI) to a JSON object. 
Each concept described in \cref{sec:design} is assigned a unique IRI in our JSON-LD models, e.g., \inlinecode{bdd:FluentClause} in \cref{lst:example-jsonld}.
Instances, i.e. models, of these concepts are represented as nodes in the JSON-LD graph, marked by the \inlinecode{@graph} keyword.
An instance links to its \mms{} via the \inlinecode{@type} keyword, e.g., \inlinecode{pickup-then} is of type \inlinecode{bdd:Then} (highlighted red).
A JSON-LD node can have multiple types, indicated by an array of IRIs.
An edge between nodes in a JSON-LD graph is declared as an object of type \inlinecode{@id}, which denotes that its value should be an IRI, i.e. a symbolic pointer to a node in the graph.
This mechanism represents \emph{relations} in our models, e.g., \inlinecode{clause-of} in \cref{lst:example-jsonld} is the relation between nodes \inlinecode{flc-held-pck} and \inlinecode{pickup-then}.

Several features of JSON-LD motivate our use of the standard to represent our models.
The mechanism to identify entities using IRIs in JSON-LD (and other RDF realizations) allows introducing additional knowledge about an entity without changing the original artefact.
This simplifies the extension of models with additional types and relations as needed, essential for managing the complexity of robotic domains (\textbf{C2}).
Furthermore, a symbolic pointer in JSON-LD only require its target to be a valid IRI, which decouples a model's structural specification from its specific realization.
This allows such realizations to be driven by their specific application context, minimizing limitations on future use of the models.
These features align well with the open-world assumption, where unforeseen extensions and applications of models are anticipated.
This is one of the major principles for designing composable models discussed in~\cite{schneider2023}, where they are applied to develop a \mm{} for representing kinematic chains and their solver algorithms.

\begin{lstlisting}[
  float,
  floatplacement=htbp,
  language=sparql,
  label=lst:example-sparql,
  caption={
    Example graph query in SPARQL.
    The \texttt{WHERE} clause specifies matching patterns to find in the graph, whereas \texttt{CONSTRUCT} creates a graph from the results.
  }
]
CONSTRUCT {
  %*{\color{mmscenario}?scenario*) bdd:has-clause %*{\color{mmfluent}?fluentClause*) .
  %*{\color{mmfluent}?fluentClause*) bdd:holds-at %*{\color{mmtc}?timeConstraint*) .
}
WHERE {
  %*{\color{mmscrtmpl}?scenarioTmpl*) bdd:of-scenario %*{\color{mmscenario}?scenario*) ;
                bdd:has-clause %*{\color{mmfluent}?fluentClause*) .
  %*{\color{mmfluent}?fluentClause*) a %*{\color{mmfluent}?fluentType*) ;
          bdd:holds-at %*{\color{mmtc}?timeConstraint*) .
}
\end{lstlisting}

JSON-LD's conformance with RDF also allows adoption of standards and tooling from its vast ecosystem.
For instance, we use SPARQL\footnote{\url{https://www.w3.org/TR/sparql11-query/}} to query and manipulate our models, extracting relevant information and constructing preferred representations from the query results, such as shown in \cref{lst:example-sparql}.
This enables a wide range of model-to-model or model-to-text transformations.
Furthermore, mechanisms for adding or modifying a graph can also be exploited for adapting models at design time or runtime~\cite{schneider2023}.
Adaptation, however, is beyond the scope of this paper.
A SPARQL query can be considered an implicit structural constraint on a model, as querying an invalid model would not yield valid results.
Explicit structural constraints on an RDF graph, e.g., types and cardinality shown in \cref{fig:bdd-overall}, can be specified using the Shapes Constraint Language (SHACL)\footnote{\url{https://www.w3.org/TR/shacl/}}, which includes more optimized mechanisms for verifying graph structures.

\begin{lstlisting}[
    float,
    floatplacement=tp,
    language=textxbdd,
    label=lst:example-textxbdd,
    caption={
        Sample BDD model in \libbddtextx{}.
    }
]
import "lab.scene"

Event pickup_start
...
%*{\color{mmscrtmpl}\textbf{Scenario Template}*) (ns=bdd_tmpl) %*{\color{mmscrtmpl}tmpl\_pickplace*) {
  %*{\color{mmvar}\textbf{var} target\_object*)
  ...
  Given:
    <%*{\color{mmvar}target\_object*)> %*{\color{mmfluent}is located at*) <%*{\color{mmvar}pick\_ws*)>
                    %*{\color{mmtc}before*) <pickup_start>
  When:
    <%*{\color{mmvar}agent*)> %*{\color{mmbhv}picks*) <%*{\color{mmvar}target\_object*)>
            %*{\color{mmbhv}and places it at*) <%*{\color{mmvar}place\_ws*)>
  Then:
    <%*{\color{mmvar}agent*)> %*{\color{mmfluent}does not collide*) <%*{\color{mmvar}place\_ws*)>
            %*{\color{mmtc}from*) <pick_start> %*{\color{mmtc}until*) <place_end>
    <%*{\color{mmvar}target\_object*)> %*{\color{mmfluent}is located at*) <%*{\color{mmvar}place\_ws*)>
                    %*{\color{mmtc}after*) <place_end>
}
%*{\color{mmus}\textbf{User Story}*) (ns=bdd) %*{\color{mmus}us\_pickplace\_lab*) {
  Scenarios:
    %*{\color{mmscrvariant}\textbf{Scenario Variant} scr\_panda\_pickplace*) {
      template: <%*{\color{mmscrtmpl}tmpl\_pickplace*)>
      scene: <%*{\color{mmscene}lab\_scene*)>
      <%*{\color{mmvar}target\_object*)> can be:
      - obj <%*{\color{mmenvagn}lab\_scene.cube1*)>
      ...
    }
}
\end{lstlisting}

\paragraph{\libbddtextx{}}
While JSON-LD is well suited to capture the intricacies of robotic domains, its verbosity and the inherent complexity of graph structures make it cumbersome to use directly.
To simplify the creation and utilization of our models, we develop a text-based Domain-Specific Language (DSL), named \libbddtextx{}, for specifying BDD criteria.
We design the DSL using the meta-language (language for defining languages) \libtextx{}~\cite{dejanovic2017}.
\cref{lst:example-textxbdd} shows a snippet of a \libbddtextx{} model, which includes a scenario template for the pick-and-place behaviour, and a variant with specific target objects.
A separate language is used to specify objects, agents and workspaces expected in the scene, which can be imported into the \libbddtextx{} model, e.g., \inlinecode{lab\_scene} on line 22.
\libbddtextx{} models can then be transformed into JSON-LD using \libtextx{}'s code generation mechanism.

Compared to the JSON-LD representation, the design of \libbddtextx{} involved several decisions that trade off the composability of the graph structure for a more user-friendly format.
For example, an instance in \libtextx{} can only have one type, meaning that transforming a multi-type JSON-LD node into \libbddtextx{} requires removing all but one of these types.
Furthermore, the more concise format of \libbddtextx{} necessitates concealing some of the concepts described in \cref{sec:design} from the user, e.g., the fluent clauses in \cref{lst:example-textxbdd} using a more specific, event-based time constraint that may hinder adoption of other temporal formalisms.

\section{Automated Execution of Acceptance Tests} \label{sec:exec}


\begin{figure}[tp]
    \centering
    \begin{subfigure}[b]{0.325\linewidth}
        \centering
        \includegraphics[height=0.13\textheight]{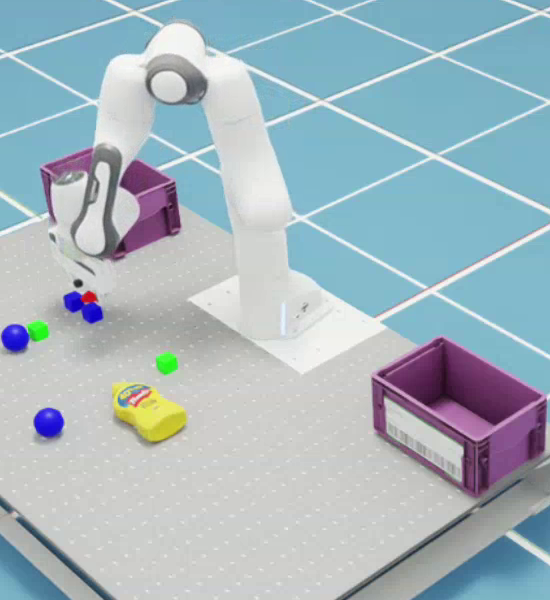}
    \end{subfigure}
    \begin{subfigure}[b]{0.325\linewidth}
        \centering
        \includegraphics[height=0.13\textheight]{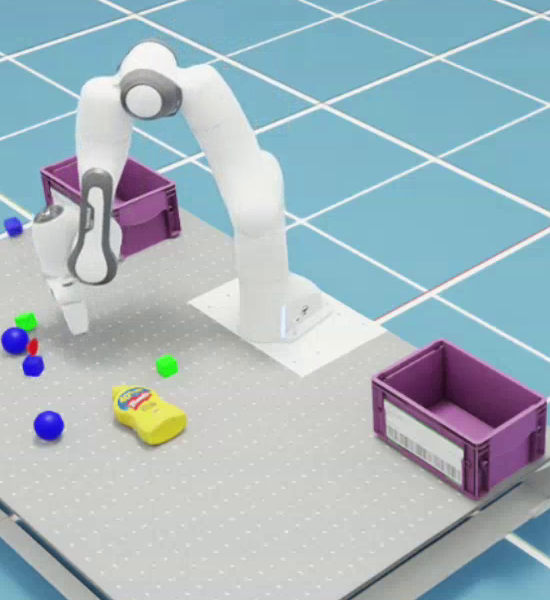}
    \end{subfigure}
    \begin{subfigure}[b]{0.325\linewidth}
        \centering
        \includegraphics[height=0.13\textheight]{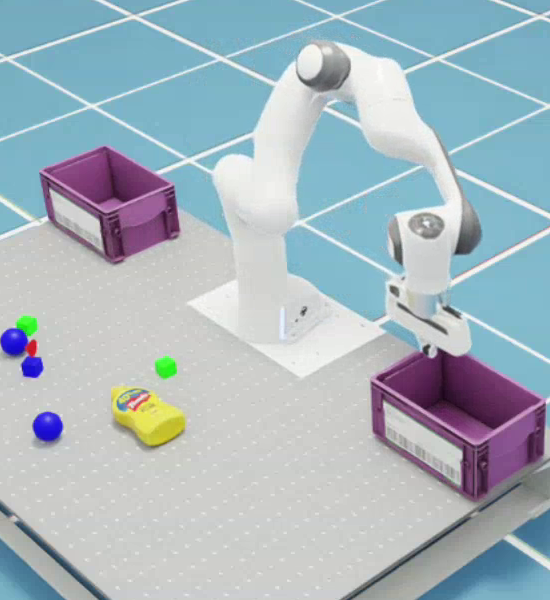}
    \end{subfigure}
    \\
    \begin{subfigure}[b]{\linewidth}
    \begin{lstlisting}[%
        language=console,%
        label=lst:console_isaac_pick_fail,
        ]
%*\color{codegreen}Given a set of objects*)
| name |
| lab:obj-cube1 |
| lab:obj-ball1 |
...
%*\color{codegreen}Given \textbf{"lab:obj-cube1"} is located at \textbf{"lab:ws-table"} before event \textbf{"tmpl:pick-start"}*)
%*\color{codegreen}When \textbf{"isaac:panda"} picks \textbf{"lab:obj-cube1"} and places it at \textbf{"lab:ws-bin2"}*)
%*\color{codegreen}Then \textbf{"isaac:panda"} does not collide \textbf{"lab:ws-bin2"}*)
      %*\color{codegreen}from \textbf{"tmpl:pick-start"} until \textbf{"tmpl:place-end"}*)
%*\color{codegreen}Then \textbf{"lab:obj-cube1"} is located at \textbf{"lab:ws-bin2"} after event \textbf{"tmpl:place-end"}*)
%*\color{mmred}AssertionError: obj 'lab:obj-cube1' (position=[...])*)
                %*\color{mmred}not located at 'lab:ws-bin2', bounds: [...]*)
    \end{lstlisting}
    \end{subfigure}
    \caption{
        Failed pickup of a cube in Isaac Sim and corresponding console output (excerpt) from \libbehave{}.
        The gripper slipped and push the target cube away before grasping.
    }
    \label{fig:bdd-isaac-pick}
\end{figure}

\noindent
In the following, we examine to what extent the acceptance testing process can be automated (\textbf{RQ2}).
To this end, we develop a prototype acceptance testing solution for a pick-place behaviour in Nvidia Isaac Sim\footnote{
The tested behaviour is in the \texttt{PickPlaceController} class of the \texttt{omni.isaac.manipulators.controllers} module (version 4.2.0).
See \url{https://developer.nvidia.com/isaac-sim}.} using the built-in implementation available for the Emika Panda\footnote{\url{https://franka.de/}} and UR10\footnote{\url{https://www.universal-robots.com/}} robotic manipulators.
Our solution includes automatic generation of Gherkin feature files with all combinations of object, workspace, and agent variations, as well as relevant domain-specific configurations to initialize the simulation.
We utilize the execution mechanism of \libbehave{}\footnote{\url{https://behave.readthedocs.io/}} (A Python library for executing Gherkin-based BDD tests) to consume the generated artefacts, execute and evaluate the pick-place behaviour in Isaac Sim.
\cref{fig:bdd-isaac-pick} shows this setup for a scenario in which the Panda robot failed to pick up an object.
This practice helps us identify areas that require further development and modelling to fully automate test execution and evaluation in simulation and on real robots.
Using the representations described in \cref{sec:dsl}, we perform the following steps to achieve automated execution of acceptance tests:



\paragraph{Generating executable BDD specifications}
To utilize existing BDD frameworks for execution, we generate Gherkin features from the JSON-LD representation using the Jinja template engine\footnote{\url{https://jinja.palletsprojects.com/}}.
Objects, workspaces, and agents associated with the \keyword{Scene} are transformed into tables under the \texttt{Background} section in the Gherkin feature file.
Variations of variables specified by the \keyword{TaskVariation} are used to generate different Gherkin \texttt{Scenario}'s.
Compared to the \texttt{Examples} table in Gherkin, \keyword{TaskVariation} is decoupled from the table data format, allowing for different variation mechanisms.
In our case, we generate variations as the Cartesian product of possible values for variables in the template.
Specifically, the robot should pick one of seven objects and place in one of the two bins, as shown in~\cref{fig:bdd-isaac-pick}, resulting in 14 scenario variations.
In each variation, variables are substituted with IRIs of objects, agents, and workspaces, whose domain-specific models, e.g., kinematics and dynamics, can be queried from the graph at runtime.
We use this mechanism to load and configure object and agent models before each scenario execution, which includes configuration variations evaluated in~\cref{sec:results}.


\paragraph{Coordinating scenario executions}
We use \libbehave{} to coordinate the execution of generated feature files.
Similar to other BDD frameworks, \libbehave{} sequentially executes matching step functions in Python for each clause in a scenario.
As BDD frameworks do not limit the implementation of step functions, we manually develop them from a robotic engineer's perspective to test the pick-place implementation provided by Isaac Sim.
Automating our BDD acceptance testing solution, e.g., via generating step implementations, requires coordinating the execution of
\begin{inparaenum}[\itshape (i)\upshape]
    \item the tested behaviour,
    \item the collection of relevant information provided by the SuT, and
    \item the evaluation of the scenario's clauses.
\end{inparaenum}

In our testing solution, the step implementation for the $IsLocated$ \texttt{Given} and \texttt{Then} clauses compare the target object's position to 3D bounds of the target workspace in the world coordinate frame, where position and bound measurements are collected during step execution.
The step function for the \texttt{When} clause executes the pick-place motion and collects the end-effector's speeds and the target workspace's positions during the behaviour.
We compute the accumulated displacement of the target bin to evaluate the $\neg collides$ \texttt{Then} clause in its step implementation, where a collision is detected if this displacement exceeds a threshold (\qty{0.05}{\metre}).

\section{Evaluating Acceptance Test Executions} \label{sec:results}
\noindent
To investigate \textbf{RQ3}, we execute variations of the pick-place implementation provided by Isaac Sim using the testing solution described in \cref{sec:exec} and evaluate the results.
Aside from variations in the target object, workspace, and agent, we introduce two additional variations in the objects' configurations:
\begin{inparaenum}[\itshape (i)\upshape]
    \item distribution ranges for sampling objects' initial position in the world frame, and
    \item height scale of the bins for placing.
\end{inparaenum}
Specifically, we sample object positions uniformly either from $ (0.25, -0.4) \le (x, y) < (0.6, 0.4) $~\si{m} or a smaller range, i.e. denser positions, $ (0.35, -0.3) \le (x, y) < (0.6, 0.3) $~\si{m}, where the ranges are chosen so that the objects are on the table in front of the robot.
Object orientations are sampled using the uniform random rotation function from SciPy.
The scale of the bins' height are either $0.8$, $0.9$, or $1.0$.
Such configurations of scenario elements are currently loaded from the RDF graph during test execution and not part of scenario specification, but we plan to incorporate them into \libbddtextx{} and the Gherkin generation process in the future.

\begin{figure}[tp]
    \centering
    \begin{subfigure}[T]{0.52\linewidth}
        \includegraphics[width=\linewidth]{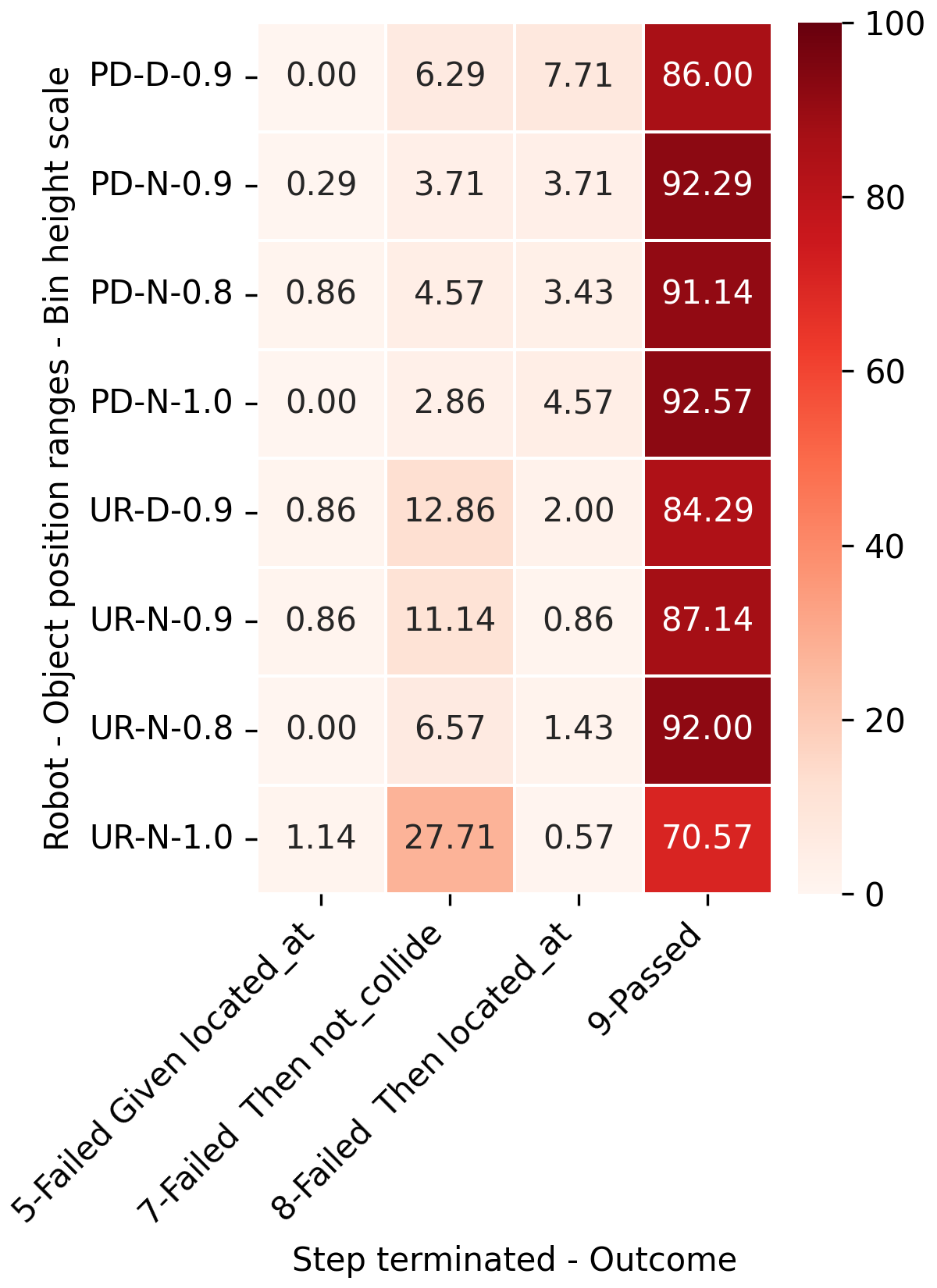}
        \subcaption{Heat map for percentage of execution outcomes (PD: Panda, UR: UR10, D: Dense, N: Normal).}
        \label{fig:results:heatmap}
    \end{subfigure}
    \hfill
    \begin{subfigure}[T]{0.46\linewidth}
        \includegraphics[width=\linewidth]{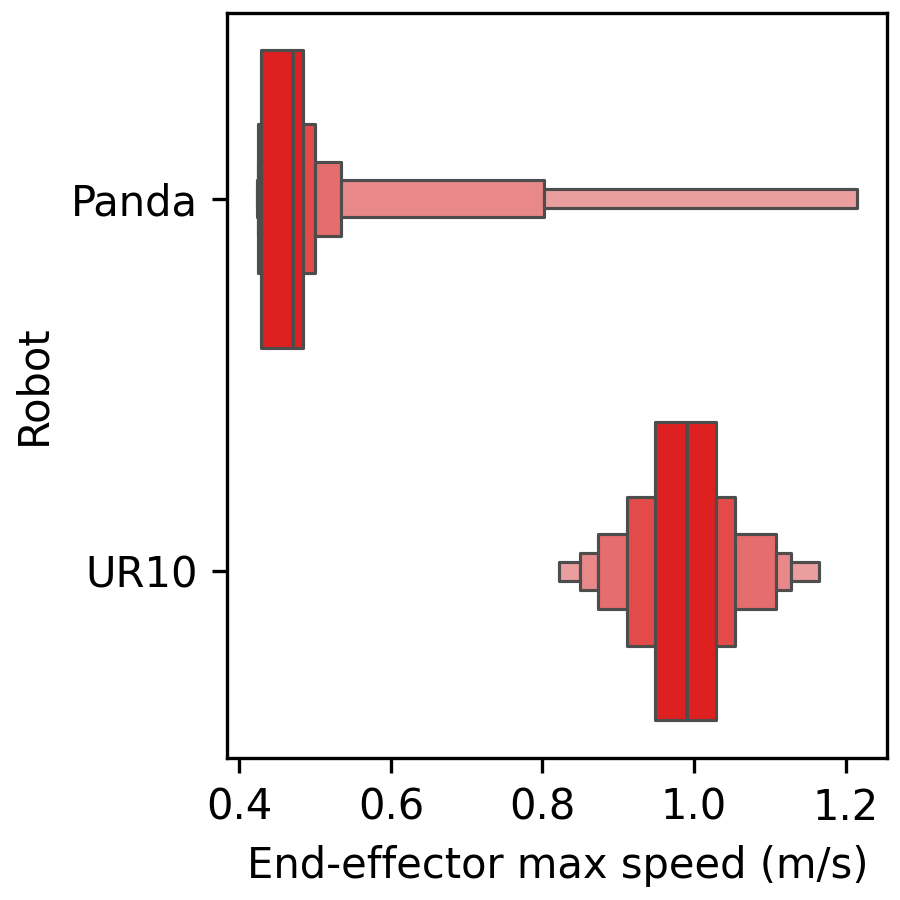}
        \vspace{11pt}
        \subcaption{
            Distributions of max EE speeds in 427 passing scenario executions with the Panda robot and 329 with the UR10, where the boxes on each tail represent the distributions' quartiles, octiles, and so on.
        }
        \label{fig:results:speed}
    \end{subfigure}
    \caption{
        Statistics for 140 executions per variation of robot, object position ranges, and placing bin height scale.
    }
    \label{fig:results}
\end{figure}

For each combination of robot, position range, and bin height scale, the generated Gherkin feature was executed ten times, resulting in 140 scenario executions per combination.
\cref{fig:results:heatmap} illustrates the percentage of execution outcomes, wherein an outcome can be categorised as a scenario passing or failing in a particular clause.
The \texttt{Given} clauses for scene setup and the \texttt{When} clause for the pick-place behaviour never fail and are omitted.
Overall, the Panda robot was more likely to fail when the objects were more densely positioned, whereas the UR10 robot failed more often when the bin was taller.
Most UR10 failures occurred in the $\neg collides$ clause, resulting from displacement of the placing bin.
The issue arises because the built-in pick-place function in Isaac Sim, after grasping an object, moves it at a fixed height of \qty{0.3}{\meter} in the world frame before placing.
This setting, which equals the end-effector (EE) height prior to picking, is not explained in the code comments or tutorial.
For the UR10 robot, which employs a suction cup as its EE by default, this can lead to collisions between the suspended object and the bin during transport before placing, particularly if the bin is tall or the object is large.
In comparison, the Panda robot is equipped with a parallel gripper and is less affected by this parameter.
Instead, most of its failures are due to the gripper slipping on different surfaces during grasping, causing the arm to reach abnormal speeds and/or the gripper failing to grasp the target object, as shown in \cref{fig:bdd-isaac-pick}.
Such slippage is more frequent when objects are closely packed, and the resulting erratic movements are reflected in the distributions of maximum EE speeds in successful scenarios, as shown in \cref{fig:results:speed}.
While the UR10's EE max speeds approximate a normal distribution around \qty[per-mode=symbol]{1}{\metre \per \second}, the Panda's EE max speeds can vary over a much wider range above its typical values of about \qty[per-mode=symbol]{0.5}{\metre \per \second}.

\section{Discussion} \label{sec:discussion}
\noindent

In our experience with developing the BDD framework, utilising knowledge graphs to represent robotic acceptance criteria facilitates a straightforward mechanism for domain-specific extensions, e.g., temporal and behaviour models.
Through code generation and semi-automated test execution, we gathered metrics from numerous scenario executions to reason about the SuT's behaviour and failure causes.
The explicit representation of time constraints is also an opportunity to further automate the coordination of test execution.

Nevertheless, further investigation is needed to better leverage these models for automated test execution.
First, time constraints on fluents and associated data collections may span only part of the behaviour execution or overlap with other fluents.
For example, while the $isHeldBy$ fluent should hold after grasping and before picking, $\neg collide$ should hold true throughout the behaviour.
Step implementations must consider such time-dependence while coordinating data collection and evaluation for these fluents, which is challenging with current BDD frameworks.
Furthermore, testers should be able to choose how to evaluate acceptance criteria depending on specific application context and system capability.
In our implementation, we use the metrics that are freely available from Isaac Sim, e.g. object position.
On a real robot, additional components would be required to check if the object is in the placing bin, e.g., recognition using visual data.
Managing such complexity and variability in evaluating robotic acceptance criteria motivates modelling effort for specifying the measurements an SuT can provide.
\section{Related Work} \label{sec:related}
\noindent
Although software testing is extensively studied~\cite{actesting}, robotic systems require specialized testing methods, concepts, and tools~\cite{vvsurvey}.
Afzal \emph{et al.}~\cite{afzal} identified certain practices established in robotics, e.g., field-testing, but others present challenges, e.g. standard conformance testing~\cite{sohail2023}.
Notably, acceptance testing in robotics remains underexplored.
Zheng \emph{et al.}~\cite{zheng} proposed using BDD to generate runtime monitors to verify timing properties in metric temporal logic, addressing one of our identified deficits.
While their monitors target Java code instrumentation, our approach aims for higher test automation through simulation environments.
Constructing relevant scenarios in simulation-based robotic tests is challenging~\cite{simsurvey}, which we addressed by using knowledge graphs to represent robotic acceptance criteria, enabling various transformation and automation activities.
Alferez \emph{et al.}~\cite{Alferez2019} proposed a model-based approach to generate BDD criteria for common data manipulation behaviours in the financial domain, but lacks robotic-specific BDD extensions.
Dos Santos \emph{et al.}~\cite{dossantos2024AAT4IRS} applied mutation testing to evaluate acceptance test executions of a robotic task in simulation, but specified the Gherkin features manually.


\section{Conclusion and Future Work}
\label{sec:conclusion}
\noindent
We introduced a model-based approach for specifying robotic acceptance tests using BDD, addressing key deficits in existing BDD methods, especially in specifying timing constraints and representing interdependent scenario relations.
To our knowledge, this is the first model-based acceptance testing approach for robots.
We formalized our model using a machine-readable knowledge graph and a DSL for robotic developers to specify BDD scenarios.
Using these models, we developed a solution for conducting acceptance tests for a pick and place task in simulation.
Through automation and execution of these tests, we identified BDD framework limitations in coordinating scenario execution and noted that further modeling is needed to represent measurements and estimations for automated scenario evaluation.
Future investigations will focus on these topics to enable automated benchmarking of robotic applications, such as in competitions and evaluation campaigns.

\printbibliography

\end{document}